\documentclass{llncs}

\usepackage{hyperref}
\usepackage{url}

\usepackage{pgf}
\usepackage{amsmath}
\usepackage{amssymb}
\usepackage{multirow}
\newcommand{\comment}[1]{}

\title{Compressing Word Embeddings}

\author{Martin Andrews}
\authorrunning{Martin Andrews} 
\institute{
  Red Cat Labs, Singapore    \\
  \email{Martin.Andrews@RedCatLabs.com} \\ 
  \url{http://www.RedCatLabs.com}
}

\begin{document}

\maketitle

\begin{abstract}
Recent methods for learning vector space representations of words have succeeded in capturing 
fine-grained semantic and syntactic regularities using large-scale unlabled text analysis.  
However, these representations typically consist of dense vectors 
that require a great deal of storage and cause the internal structure of the vector space to be opaque.  
A more `idealized' representation of a vocabulary would be both compact and readily interpretable.
With this goal, this paper first shows that Lloyd's algorithm can compress the standard dense vector 
representation by a factor of 10 without much loss in performance.
Then, using that compressed size as a `storage budget', we describe a new GPU-friendly factorization 
procedure to obtain a representation which gains interpretability as a side-effect of 
being sparse and non-negative in each encoding dimension.
Word similarity and word-analogy tests are used to demonstrate the effectiveness of the compressed
representations obtained.

%
%
%

%
%
%
%
%
%
%

%

\end{abstract}

\section{Introduction}

Distributed  representations  of  words  have  been shown to benefit 
NLP tasks like parsing, named entity recognition, and sentiment analysis, 
as well as being used as the raw material for other deep learning tasks.   

Surprisingly, these word vector embeddings can be derived directly from raw, unannotated corpora.
Once created, the vector embedding $E$ can be expressed simply as a 
list of vocabulary words (of size $V$), and a matrix of size ${V}\times d$, 
where $d$ is the dimensionality of the embedding space.


As argued in Murphy et al. (2012)\cite{murphy2012learning}, 
while this dense matrix representation may be handled with ease by computers, 
there are cognitive arguments against such a representation being the basis of language 
(see Griffiths et al. (2007)\cite{griffiths2007topics} for broader discussion on this point). 
For instance, it seems unlikely that the same small set of features are sufficient and necessary to describe all
semantic domains of a full adult vocabulary. 
%
It would also be uneconomical for people to store all negative properties of a concept, 
such as the fact that dogs do not have wheels, or that airplanes are not used for communication. 
Indeed, in feature norming exercises (for example Vinson and Vigliocco, 2008 \cite{Vinson2008}) 
where participants are asked to list the properties of a word, 
the aggregate descriptions are typically limited to approximately 10-20 characteristics for a given concrete concept.

So, for cognitive plausibility, we claim that a feature set should have three characteristics: 
it should only store positive facts; 
it should have a wide range of feature types, to cover all semantic domains in the typical mental lexicon; 
and only a small number of these should be active to describe each word/concept.

\section{Models}

In this work, a 300-dimensional GloVe embedding \cite{pennington2014glove} 
was used as a concrete baseline for a number of different lossy compression methods
\footnote{The techniques described in this paper can be applied to any embedding, 
since nothing specific to GloVe has been used.}.

\comment{

GloVe embeddings are produced by minimising the total error $|J|$ in predicting 
the word co-occurrence matrix $X_{i,j}$ from the word vectors associated with words $i$ and $j$: 

\[ J = \sum_{i,j=1}^{V} f(X_{ij}) ( w^T_i \tilde{w}_j + b_i + \tilde{b}_j - \log{X_{ij}} )^2 \]

where $f()$ is a function that (i) vanishes at zero, to eliminate problems of $X_{ij}$ being zero in the $\log()$ term, 
and (ii) levels off once $X_{ij}$ is `large', so that frequent words don't dominate the error calculation.

Once the $w_i$ and $\tilde{w}_j$ have been determined (the `word vectors' and `context vectors' respectively),
the embedding matrix $E$ is set so that row $e_i = w_i$. 
\footnote{Alternatively, $e_i = (w_i + \tilde{w}_i)$ could have been used, but it made little difference in practice.}

}




\subsection{Simple Compression}
A number of different compression methods were explored to establish a `storage budget' for 
the subsequent experiments in sparse encodings.  These intial methods focussed on
discarding data, while resulting in an approximately equivalent level of performance in
the `Google' word analogy task (described later).

\comment{
Figure \ref{figure:normal} illustrates the significant non-normality of the distributions of the values in each 
vector dimension across all vocabulary words.
}

The methods that were explored included (i) directly discarding a fixed proportion of the vector dimensions;
(ii) thresholding the data; (iii) quantising the data based on curves of the form $\pm|v_i|^\alpha$ for each dimension
$i$ of a given vector $v$; and (iv) adaptive level encoding.  
This last approach, efficiently implemented via Lloyd's algorithm \cite{Lloyd:2006:LSQ:2263356.2269955}, 
was shown to be capable of approximating each element $v_i$ of the entire vocabulary with only
8 discrete levels (each dimension $i$ being independently calibrated) while still performing
acceptably on the analogy task.

Thus, for the 300-dimensional embedding, a 900-bit `storage budget' was established, and
used as the storage bound for the sparsification experiments.


\comment{

\begin{figure}[ht]
 \caption{Normal Plots of the embedding element distributions}
 \begin{center}
  \input{figure_normal-plot.pgf}
 \end{center}
 \label{figure:normal}
\end{figure}

}

\subsection{Lloyd's algorithm}

For a given number of quantisation levels $n$ and an element $e\sim E$, 
we want to create a set of quantisation levels $e_q$ that minimises:
\[ \sum_{e\in{E}}{ \min_q{(e - e_q)^2} } \]

This process can be performed iteratively, starting with the $e_q$ placed uniformly 
within the values of $E$ sorted numerically.  

The algorithm then iteratively updates the $e_q$ such that the centroid of each neighbouring cluster is found, 
and the $e_q$ are then updated to these positions.  The algorithm stops when no more changes are required.


This optimisation is done for each element/column of the embedding independently, 
and a list of the respective levels of $e_q$ is stored, in addition to the index within each list for each
element of the representation.

\subsection{Sparse Compression}


\subsubsection{Non-Negative Sparse Embedding (``NNSE'')}
A very promising approach to sparse embedding was taken in Murphy et al. (2012)\cite{murphy2012learning},
where the following optimisation was solved iteratively to generate a sparse embedding $A$ (which is $V$ vectors each with dimensionality $k$),
and also creates a `dictionary' $D$ (which maps the sparse elements of $A$ onto the real-valued embedding $E$) : 
\begin{multline*}
\begin{split}
argmin_{ A\in\mathbb{R}^{m\times{k}}, D\in\mathbb{R}^{k\times{n}} } 
& \sum_{i=1}^{V}{ ||E_{i,:} - A_{i,:} \times D||_2 + \lambda||A_{i,:}||_1 } \\
\text{where : } &  D_{i,:}, D^T_{i,:} \leq 1, 1 \leq \forall i \leq k \\
\text{and } & A_{i,j} \geq 0,  1 \leq \forall i \leq V, 1 \leq \forall j \leq k
\end{split}
\end{multline*}

Although this produced excellent embedding results, one requirement is to vary the parameters
$k$ and $\lambda$ such that the sparse embedding $A$ had a `reasonable' number of positive entries.
Since one of the targets of this paper is compression, targetting a concrete compression level,
this hyperparameter search is not acceptable.

\subsubsection{Winner-Take-All Autoencoders}

Another approach to sparse encoding, applied chiefly to visual tasks, was described in Makhzani and Frey (2014\cite{DBLP:journals/corr/MakhzaniF14}).
Their ``Winner-Take-All Autoencoders'' obtained strict sparsity targets by interposing 
a drop-out layer that only allows the top-$\alpha$ fraction of the layer's inputs to be passed through, while the 
rest are set to zero.  
An important implementation detail is that this drop-out is done on a \emph{per dimension} basis over each minibatch of training examples.
This implicitly causes the training to balance the occurrences of non-zero entries across the different dimensions
of the embedding space, while allowing the $A_{i,:}$ vectors to have varying numbers of non-zero entries.

Unfortunately, obtaining the top-$\alpha$ fraction of a set of numbers requires sorting the list, 
which means that including this layer in an neural network will move data-intensive computations off the GPU.
In order to produce a sparse embedding in reasonable time, however, the GPU is required, 
so an algorithmic short-cut was implemented that can be efficiently executed inside the GPU.

\subsubsection{GPU-friendly top-$\alpha$}

Clearly, determining the top-$\alpha$ percentile of a distribution is equivalent to determining the hurdle value
$h^\ast$ above which the fraction $\alpha$ of the realised values in the minibatch lie
\footnote{A minibatch has 16,384 examples -- large enough for distribution approximations.}.

For a given embedding dimension $j$, and assuming that $\alpha\ll\text{50\%}$ and that the distribution is not pathelogical, 
this hurdle can be bounded above by 
$h^{+}_j = max(A_{:,j})$ 
and below by 
$h^{-}_j = mean(A_{:,j})$.  A binary section search can then be made for $h^\ast$, since the fractional percentile
that any given $h$ corresponds to can be determined by taking the mean of a simple matrix $A_{:,j}>h$ indicator function.
\footnote{Also, $\alpha^{+}_0=(1/\text{batchsize})$ initially, since it is the maximum value in $A_{:,j}$}.

Although a thorough implementation could include stopping criteria, 
that would not make sense within the GPU's cores' execution flow.  
So a fixed number of bisection iterations was used (specifically $5$ for our 16,384 minibatch size).  
This was found to give robust results, with a value of $h$ being found 
that identified (approximately) the top-$\alpha$ percentile consistently.
The net result of avoiding CPU operations was a $39\times$ speed-up.

Experiments were also conducted to try and optimise the placement of each $h$ estimate, using distributional
properties of $N()$, or linear interpolation.  These were not successful, however, since it 
appears that the neural network learns to exploit the distribution assumptions being made, so 
as to win artificially low $h$ (and thus a higher accepted $\alpha$ than required).

\subsubsection{Sparse Autoencoding for Word Embeddings}

%

To avoid learning \emph{from scratch} the entire sparse matrix $A$, an autoencoding scheme 
was set up, so that the target embedding $E$ was mapped onto intermediate layers, then
quantised, then mapped back to itself (via a trainable dictionary $D$).

The layers of the implemented autoencoder are described in Table \ref{table:quasiauto}, where
concrete sizes are given, assuming a vocabulary size of $2^{17} = 131,072$ words, $300$-dimensional
underlying embedding, and a sparse embedding of size $k=1024$.

The quantity being optimised via gradient descent was the $l_2$ error between the output embedding $E^\ast$,
and the input $E$, and this scheme was accelerated using ADAM (Kingma et al., 2014\cite{kingma2014adam}).

Note that no regularisation terms were used.  There is regularisation being implicitly applied due to the 
top-$\alpha$ sparsity constraint, coupled with the batch normalization that occurs in the layer before the
sparse encoding.  These combine to constrain the sparse encoding $A$ within reasonable bounds.

In order to improve convergence, a scheme was used whereby a variable $\sigma$ (initially zero) was 
incremented (by $0.01$) after each epoch \emph{only if} the $l_2$ error over the epoch averaged less than $0.01$.
The variables $\alpha_t$ and $\beta_t$ were then dynamically adjusted as follows:
\begin{multline*}
\begin{split}
  \alpha_t & = 50\% \times (e^{10\sigma}) + \alpha  \times (1-e^{10\sigma}) \\
  \beta_t & = 0.2 \times e^{0.01\sigma}
\end{split}
\end{multline*}

Having this path-dependent speed regulator enabled learning pressure to be applied incrementally through 
training in a way that was sensitive to current training progress.  In particular, the sparsity factor $\alpha_0$ was 
started out at `easy' values, so that the initial weights could migrate to areas where they were at least
tackling the underlying problem - and $\alpha_t$ then moved asymptotically closer to $\alpha$ over the course
of a thousand epochs.  Similarly, the Guassian noise layer seemed to improve search in later periods,
and this was decayed more slowly.

After training, the noise-layer was switched off, and the network-learned $A$ and $E^\ast$ were used for the
performance analyses below.

{
\renewcommand{\arraystretch}{1.5}
\begin{table}[h]
 \centering
 \caption{Sparse Auto-encoding model structure}
 \begin{tabular}{  l l | l }
  Parameter Set                      & Intra-layer Operations                        & Output Shape \\
  \hline 
  Word Embedding (all tokens)        & $\mathbf{E}$                                  & $(2^{17}, 300) \in\mathbb{R}$ \\
  \hline
  Hidden layer                       & $\max(\mathbf{W}^H\mathbf{x} + \mathbf{b}^H, 0)$  & $(2^{17}, 300\times 8) \in\mathbb{R}^+$ \\
  \hline
  Pre-Binary linear layer            & $\mathbf{W}^L \mathbf{x} + \mathbf{b}^L$      & $(2^{17}, 1024) \in\mathbb{R}$ \\
  \hline
  Batch Normalization Layer          & $\text{batchnormalize}(\mathbf{x})$           & $(2^{17}, 1024) \in\mathbb{R}$ \\
  \hline
  Rectification                      & $\text{max}(\mathbf{x},0)$                    & $(2^{17}, 1024) \in\mathbb{R}^+$ \\
  \hline
  Gaussian Noise                     & $\mathbf{x} + N(0, \beta_t)$                 & $(2^{17}, 1024) \in\mathbb{R}$ \\
  \hline
  Top-$\alpha$ sparsification        & $\text{drop-all-but-}\alpha(\mathbf{x}, \alpha_t)$  & $(2^{17}, 1024) \in\mathbb{R}^+$ \\
  \hline
  Sparse encoded version             & $\mathbf{A} = \mathbf{x}$                     & $(2^{17}, 1024) \in\mathbb{R}$ \\
  \hline
  Output (= $\mathbf{E^\ast}$)       & $\mathbf{W}^F \mathbf{x} + \mathbf{b}^F$      & $(2^{17}, 300) \in\mathbb{R}$ \\
  \hline
 \end{tabular}
 \label{table:quasiauto}
\end{table}
}

\subsubsection{Representation of Sparse Encodings}

Suppose a non-negative sparse encoding over a $k$-dimensional space is required, and a specific `bit budget' of $n$ bits is given.
Since the majority of encoding values will be zero, the representation need only store the 
addresses of the non-zero elements, and their values.  
And since the ordering of the list of values is an additional degree of freedom, 
one can increase the information content by storing 
(a) the length of the list, 
(b) the locations of each of the values in declining numerical order, 
(c) the highest value, and 
(d) the percentage ratio between each sucessive pair (typically in the range $[70\% ... 100\%]$).
While (c) might be stored with some fidelity, the remaining ratios are less exacting.  
Supposing that 3-bits is sufficient for each ratio (d), then the required sparsity ratio $\alpha$ can 
be determined \footnote{This is how the $\alpha$ values are chosen for the the experiments} :
%
\[ \alpha = {n \over{ k ( \log_2(k) + 3 ) } } \]

In addition to the $n\times{V}$ bits of storage required, there is the additional overhead of storing the 
dictionary $D$ - which consists of $k\times{d}$ elements, which can be a significant factor if $k$ is large.

\section{Experiments}
\label{experiments}

\subsection{Corpus}

The text corpus used here was the concatenation of 
(a) the `One Billion Word Benchmark' (Chelba et al. (2013)\cite{DBLP:journals/corr/ChelbaMSGBK13}), 
and (b) a cleaned version of English Wikipedia (August 2013 dump, using only pages with greater than 20 pageviews),
which was pre-processed by removing non-textual elements, sentence splitting, and tokenization. 
After preprocessing the corpus contained 1.2 billion tokens (680 million and 524 million from each respective source)
\footnote{Importantly, these resources have been made freely available without restrictive licenses, 
and in the same spirit, the code for this paper is being released under a permissive license.}.

Models were derived using windows of 5 tokens to each side of the focus word. 
Rather than focus on a word occurrence limit (as is common), a vocabulary-size of $2^{17} = 131,072$ words was chosen, 
since (a) it made batching more convenient, and (b) the limit is arbitrary either way.

\subsection{Test Datasets}
We evaluated each word representation on seven datasets covering similarity and analogy tasks,
using the test framework of Levy et al. (2015)\cite{levy2015improving}, 
which has code and data available at : \url{https://bitbucket.org/omerlevy/hyperwords}.

\subsubsection{Word Similarity}
Five datasets were used to evaluate word similarity: 
the popular WordSim353 (Finkelstein et al., (2002)\cite{Finkelstein2002}) partitioned into two datasets, 
WordSim Similarity and WordSim Relatedness (Zesch et al., (2008)\cite{Zesch:2008:UWC:1620163.1620206}; Agirre et al. (2009)\cite{Agirre:2009:SSR:1620754.1620758});
Bruni et al.'s (2012)\cite{Bruni:2012:DST:2390524.2390544} MEN dataset; 
Radinsky et al.'s (2011)\cite{Radinsky:2011:WTC:1963405.1963455} Mechanical Turk dataset; 
and Luong et al.'s (2013)\cite{Luong-etal:conll13:morpho} Rare Words dataset.
All these datasets contain word pairs together with human-assigned similarity scores. 
The word vectors are evaluated by ranking the pairs according to their cosine similarities, 
and measuring the correlation (Spearman's $\rho$) with the human ratings.

\subsubsection{Word Analogy}
The two analogy datasets present questions of the form ``$a$ is to $a^\ast$ as $b$ is to $b^\ast$'', 
where $b^*$ is hidden, and must be guessed from the entire vocabulary. 
MSR's analogy dataset (Mikolov et al. (2013a)\cite{mikolov2013msr}) contains 8000 morpho-syntactic analogy questions, 
such as ``good is to best as smart is to smartest''. 
Google's analogy dataset (Mikolov et al. (2013b)\cite{mikolov2013efficient}) contains 19544 questions, 
about half of the same kind as in MSR (syntactic analogies), and another half of a more semantic nature, 
such as capital cities (``Paris is to France as Tokyo is to Japan''). 
After filtering questions involving out-of-vocabulary words, i.e. words that did not appear in the pruned Corpus, 
we remain with 7118 instances in MSR and 19296 instances in Google. 

As in Levy and Goldberg (2014)\cite{levy2014linguistic}, the analogy questions are answered 
using both \textsc{3CosAdd} as well as \textsc{3CosMul}.

\comment{
(addition and subtraction), setting $VW^\ast = VW/\{a^\ast,b,a\}$:

\begin{multline*}
argmax_{b^\ast \in VW^\ast} cos(b^\ast, a^\ast - a + b) = \\
argmax_{b^\ast \in VW^\ast} (cos(b^\ast, a^\ast) - cos(b^\ast, a) + cos(b^\ast, b))
\end{multline*}

as well as \textsc{3CosMul}, which was also effective in analogy recovery :

\[  argmax_{b^\ast \in VW^\ast}  cos(b^\ast, a^\ast) · cos(b^\ast, b) / {cos(b^\ast, a) + \epsilon} \]

$\epsilon$ = 0.001 is used to prevent division by zero. 

We abbreviate the two methods ``Add'' and ``Mul'', respectively.
The evaluation metric for the analogy questions is the percentage of questions for which
the argmax result was the correct answer ($b^∗$).
}

\subsection{Results}

Embeddings with representations compressed by the Lloyd algorithm and the non-negative sparse encoding 
methods outlined above were run, with results shown in Tables \ref{table:similarity} and \ref{table:analogy}.

In addition to the reconstructed embeddings $E^\ast$ for each method, the \emph{raw results} 
of the similarity and analogy tests were run on the intermediate sparse embeddings $A$ themselves.

{
\renewcommand{\arraystretch}{1.5}
\begin{table}[ht]
 \centering
 \caption{Similarity results}
 \begin{tabular}{  c | c c c c c }
  Method & WordSim & WordSim & Bruni et al. & Radinsky et al. & Luong et al.\\
         & Similarity & Relatedness & MEN  & M. Turk & Rare Words\\
  \hline
  GloVe baseline                    & 66.2\% & 52.1\% & 69.1\% & 63.2\% & 22.8\%  \\
  \hline
  Lloyd-8                           & 65.9\% & 51.5\% & 68.6\% & 62.3\% & 22.7\%  \\
  \hline
  $E^\ast(k=4096, \alpha=1.50\%)$   & 65.0\% & 50.2\% & 68.5\% & 62.7\% & 22.2\%  \\
  $E^\ast(k=1024, \alpha=6.75\%)$   & 64.7\% & 51.0\% & 67.7\% & 62.7\% & 21.9\%  \\
  \hline
  $A(k=4096, \alpha=1.50\%)$        & 63.7\% & 45.2\% & 62.4\% & 49.5\% & 18.0\%  \\
  $A(k=1024, \alpha=6.75\%)$        & 69.2\% & 51.3\% & 69.5\% & 62.1\% & 17.4\%  \\
  \hline
  LSH-900                           & 62.0\% & 45.8\% & 65.9\% & 59.2\% & 22.0\%  \\
  
 \end{tabular}
 \label{table:similarity}
\end{table}

\renewcommand{\arraystretch}{1.5}
\setlength\tabcolsep{10pt}
\begin{table}[ht]
 \centering
 \caption{Analogy results}
 \begin{tabular}{  c | c c  }
  Method & Google & MSR \\
         & Add/Mul & Add/Mul \\
  \hline
  GloVe baseline                     & 67.1\% / 68.5\% & 53.4\% / 56.6\% \\
  \hline
  Lloyd-8                            & 65.9\% / 67.4\% & 51.9\% / 54.5\% \\
  $E^\ast(k=4096, \alpha=1.50\%)$    & 62.5\% / 66.4\% & 51.8\% / 54.1\% \\
  $E^\ast(k=1024, \alpha=6.75\%)$    & 62.4\% / 62.9\% & 49.0\% / 50.1\% \\
  \hline
  $A(k=4096, \alpha=1.50\%)$         & 37.6\% / 40.8\% & 27.3\% / 29.9\% \\
  $A(k=1024, \alpha=6.75\%)$         & 52.5\% / 55.1\% & 40.5\% / 43.8\% \\
  \hline
  LSH-900                            & 53.0\% / 53.1\% & 41.2\% / 42.2\% \\
 \end{tabular}
 \label{table:analogy}
\end{table}
}

As a point of comparison, a random vector approach 
(binarised locality sensitive hashing : ``LSH'', Charikar (2002)\cite{Charikar:2002:SET:509907.509965}) was also tested, 
with the same 900-bit constraint.

\section{Discussion}
\label{discussion}


\subsection{Level quantisation vs. more sophisticated methods}

The level-quantisation approaches to compression work extremely well, 
and are relatively simple to implement.  
Assuming values would otherwise be stored as 32-bit floats,
the Lloyd method acheives very similar scores, with only 3-bits per value
(the overhead of storing the quantisation levels is only the equivalent of 
storing 8 of the original word-vectors).
However, they do not accomplish the goal of learning about the  
underlying embedding through an efficient compression algorithm.

\subsection{Performance of Sparse Embeddings}

As can be seen from Table \ref{table:similarity}, the reconstruction ($E^\ast$) results 
for both sparse embedding methods are only marginally below those of the original embeddings,
and generally better than the LSH re-representation.  This is satisfying, because it 
shows that the GPU-friendly method outlined here actually reconstructs the 
embedding without a significant loss in performance, within the same `bit budget' as
the quantisation methods.

The fact that the sparse encodings $A$ can also perform as embeddings on their own 
(without involving the learned dictionary $D$) is encouraging, 
since it implies that there is more information about the underlying language 
that can be obtained from existing word embeddings `for free'.  

Interestingly, their performance on similarity tasks is far higher than on the analogy ones
(particularly in the case of $k=4096$ which has an $\alpha$ of only $1.50\%$).
This can be understood by considering the algebra of the sparse non-negative vectors being used.
For similarity purposes, non-negative vectors can scored by the same a cosine measure that works
for more general dense, real-valued vectors.  
However, for the analogy tasks, there are implicit subtractions being done, which result in
direction vectors that are not part of the same algebra.  
This deserves further work, to see whether other operators would be able to make use of the 
sparse vector spaces' geometry more fully 
(potentially including ideas from Mahadevan and Chandar (2015)\cite{DBLP:journals/corr/MahadevanC15}).

\subsection{Interpretability of the sparse ``$A$''}

Table \ref{table:interpretability} (which lists the highest weighted words 
in each dimension that `motorbike' is also most highly weighted) clearly demonstrates 
that the $A$ sparse representation has learned something about the 
structure of the English language `for free', using only data obtained from an
embedding trained on unlabelled data itself.

{
\renewcommand{\arraystretch}{1.1}
\setlength\tabcolsep{10pt}
\begin{table}[h]
 \centering
 \caption{Top 'motorbike' dimensions}
 \begin{tabular}{  l | l }
  Model & Top words in each of first 7 dimensions \\
  \hline
  \multirow{7}{*}{GloVe baseline}                     
& lb., four-bladed, propeller, propellers, two-bladed, ... \\ 
& passerine, 1975-79, rennae, fyrstenberg, edw, coots, ... \\ 
& bancboston, oshiomhole, 30-sept, holmer, smithee, recon, ... \\ 
& http://www.nytimes.com, (888), receival, jamiat, shyi, ... \\ 
& subjunctive, purley, 11-july, broaddus, muharram, ebit, ... \\ 
& proximus, pattani, 31-feb, wgc, 30-nov, crossgen, 2,631 \\
& officership, tvcolumn, integrable, salticidae, o-157, ... \\ 
  \hline
  \multirow{7}{*}{$A(k=1024, \alpha=6.75\%)$}         
& vehicles, vehicle, cars, scrappage, car, 4x4, armored \\
& prix, races, race, laps, vettel, rikknen, sprint \\
& ski, coal, gas, taxicab, nuclear, wine, cellphone \\
& kool, electrons, pulpit, efta, gallen, gasol, birdman \\
& eric, anglo, tornadoes, rt, asteroids, dera, rim \\
& wear, trousers, dresses, jeans, wearing, worn, pants \\
& stabbed, kercher, 16-year-old, 15-year-old, 18-year-old, ... \\
 \end{tabular}
 \label{table:interpretability}
\end{table}
}


\pagebreak 

\section{Conclusion}
\label{conclusion}

The joint goals of good compression rates and cognitive plausibility are realistic and achievable.

Using the GPU-friendly sparsity Winner-Take-All Autoencoder scheme described, 
sparse, non-negative encodings have been demonstrated that combine high compression
with interpretability `for free'.

Further work will include the investigation of operators that respect the 
geometry of these sparse vectors, so that analogy tests might perform in-line with
the word similarity scores (which are purely direction-based).




\subsubsection*{Acknowledgments}


The author thanks DC Frontiers, the creators of the
data-centric service `Handshakes' (\url{http://www.handshakes.com.sg/}), 
for their willingness to support this on-going research.
DC Frontiers is the recipient of a Technology Enterprise Commercialisation 
Scheme grant from SPRING Singapore, under which this work took place.



\bibliography{embedding}
\bibliographystyle{unsrt}

\end{document}